\documentclass[conference]{IEEEtran}
\IEEEoverridecommandlockouts
\usepackage{lipsum}
\usepackage{amsmath,amssymb,amsfonts}
\usepackage{algorithmic}
\usepackage{graphicx}
\usepackage{textcomp}
\usepackage{xcolor}
\usepackage{ifpdf}
\def\BibTeX{{\rm B\kern-.05em{\sc i\kern-.025em b}\kern-.08em
    T\kern-.1667em\lower.7ex\hbox{E}\kern-.125emX}}
\usepackage{multirow}
\usepackage[utf8]{inputenc}
\usepackage{csquotes}
\usepackage[english]{babel}

\usepackage[
backend=biber,
style=numeric,
sorting=none
]{biblatex}

\begin{document}

\title{TLU-Net: A Deep Learning Approach for Automatic Steel Surface Defect Detection\\
}

\author{\IEEEauthorblockN{Praveen Damacharla}
\IEEEauthorblockA{\textit{Research Scientist} \\
\textit{KineticAI Inc.}\\
Crown Point, IN, USA \\
Praveen@KineticAI.com}
\and
\IEEEauthorblockN{Achuth Rao M. V.}
\IEEEauthorblockA{\textit{Dept. of Electrical Engineering} \\
\textit{Indian Institute of Science (IISc)}\\
Bengaluru, KA, India\\
achuthr@iisc.ac.in}

\and

\IEEEauthorblockN{Jordan Ringenberg}
\IEEEauthorblockA{\textit{Computer Science Dept.} \\
\textit{The University of Findlay}\\
Findlay, OH, USA\\
ringenberg@findlay.edu}

\and
\IEEEauthorblockN{Ahmad Y. Javaid}
\IEEEauthorblockA{\textit{EECS Department} \\
\textit{The University of Toledo}\\
Toledo, OH, USA \\
Ahmad.Javaid@Utoledo.edu}

}

\maketitle

\begin{abstract}
Visual steel surface defect detection is an essential step in steel sheet manufacturing. Several machine learning-based automated visual inspection (AVI) methods have been studied in recent years. However, most steel manufacturing industries still use manual visual inspection due to training time and inaccuracies involved with AVI methods.  Automatic steel defect detection methods could be useful in less expensive and faster quality control and feedback. But preparing the annotated training data for segmentation and classification could be a costly process. In this work, we propose to use the Transfer Learning-based U-Net (TLU-Net)  framework for steel surface defect detection. We use a U-Net architecture as the base and explore two kinds of encoders: ResNet and DenseNet. We compare these nets' performance using random initialization and the pre-trained networks trained using the ImageNet data set. The experiments are performed using Severstal data. The results demonstrate that the transfer learning performs 5\% (absolute) better than that of the random initialization in defect classification. We found that the transfer learning performs 26\% (relative) better than that of the random initialization in defect segmentation. We also found the gain of transfer learning increases as the training data decreases, and the convergence rate with transfer learning is better than that of the random initialization. 
\end{abstract}


\section{Introduction}
\begin{figure}
     \centering
     \includegraphics[width=\columnwidth,trim={0cm 0cm 0 0},clip=true]{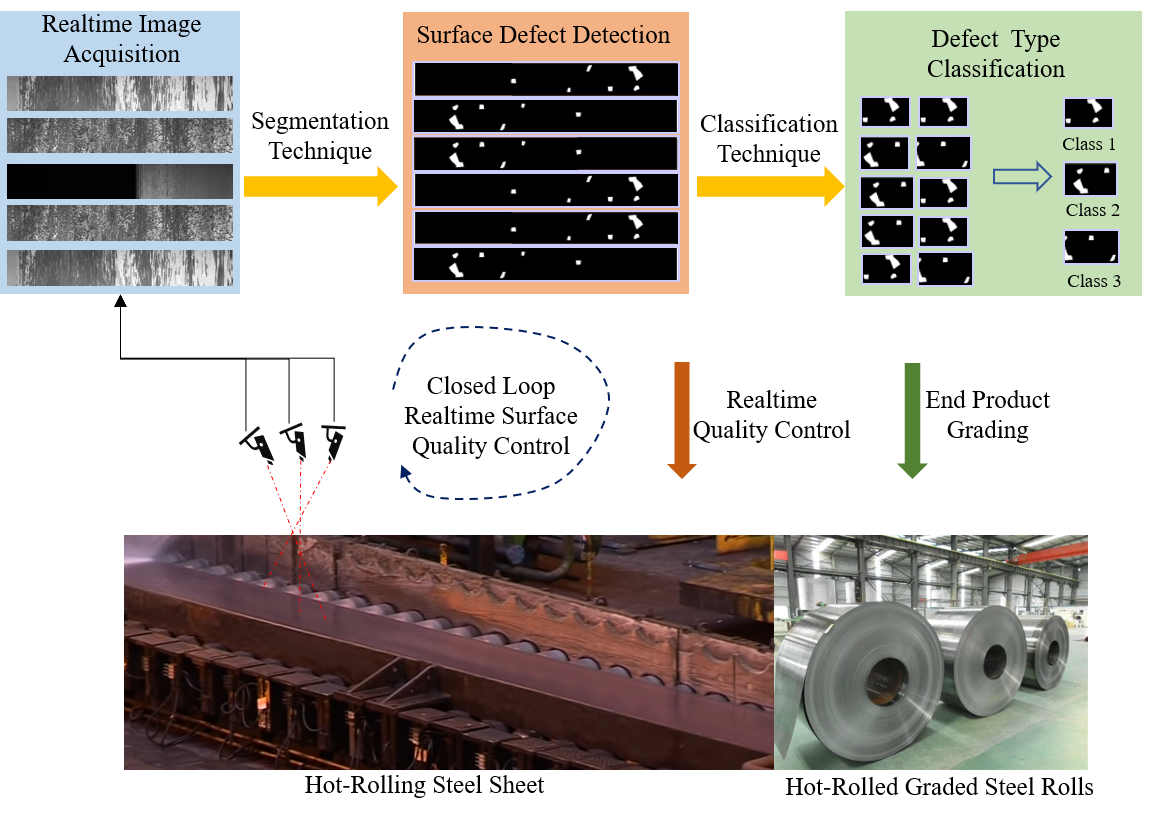}
     \caption{A Generic Automatic Visual Inspection Outline}
     \label{fig:fig1}
 \end{figure}

Steel is one of humanity's most important building materials. Defect inspection is a critical step of quality control in the steel plates. This step mainly involves capturing images of the steel surface using an industrial camera followed by recognizing, localizing, and classifying the defect, which helps rectify the defect's cause. Typically, this process is performed manually, which is not reliable and time-consuming. Unreliable quality control can cause a huge economic problem for manufacturers. Manual detection can be replaced or aided by the automatic classification using computer vision methods. The general flow of automatic visual inspection for quality control is shown in Fig. \ref{fig:fig1}.
There are two main steps involved in the defect inspection. The first step is to classify the defect type from the images, and the second step is to identify the defect location in the image. There are various automatic methods in the literature to address one/both of these steps. Some of the early methods use a handcrafted feature to classify the defect type \cite{caleb2007adaptive, song2013noise, dong2014texture}, and few methods find the coarse defect locations. The main drawback of these methods is that the features need to be designed by an experts. The designed feature may not generalize to new type of defect.  The recent advances in end-to-end deep learning (DL) methods overcame these hand-designed features. It learns to extracts the multi-scale features depending on the task using only the data and labels. The DL method has been shown to outperform the hand-designed features in various computer vision tasks \cite{lecun2015deep}.

\begin{figure*}
    \centering
    \includegraphics[width=\textwidth,trim={0cm 0cm 0 0},clip=true]{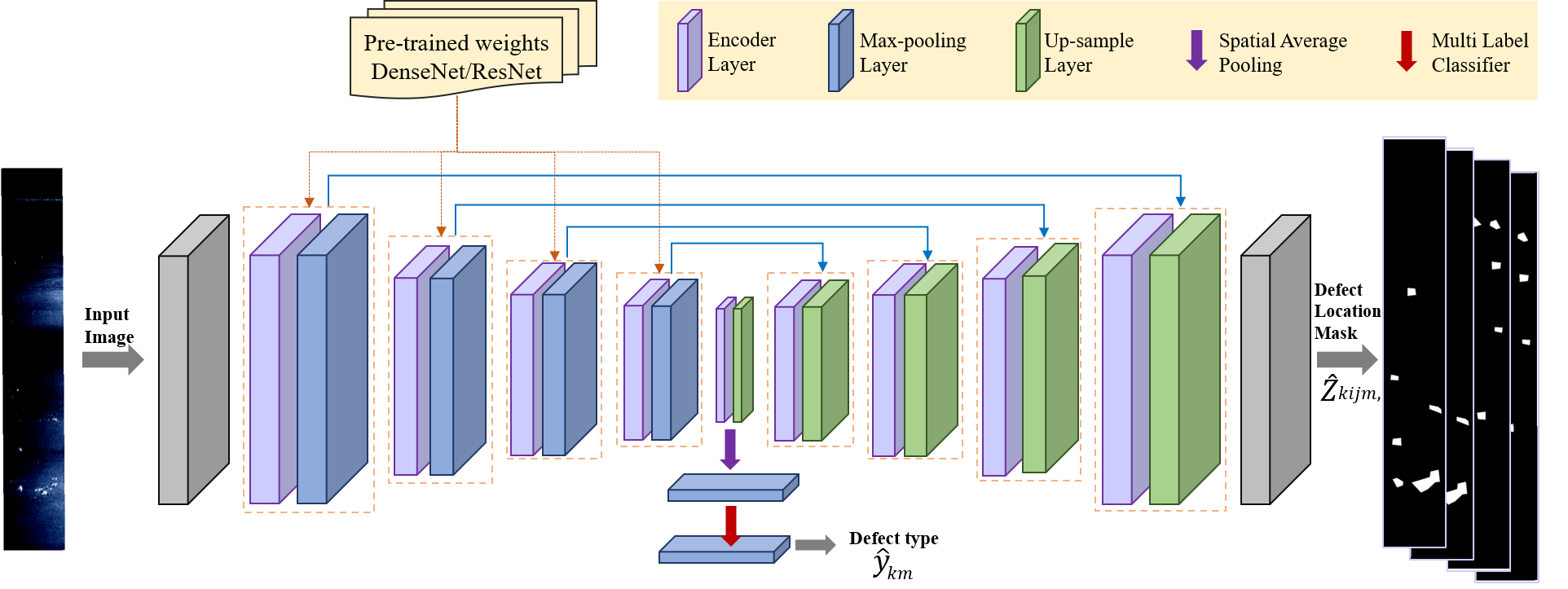}
    \caption{Proposed architecture transfer learning method for the joint steel defect classification and segmentation. The blue line indicate the skip connection and the orange dotted line indicate the initialization. }
    \label{fig:pa_arch}
\end{figure*}

There are various deep learning methods are used to perform defect classification. \cite{chen2016overfeat} use the features extracted from the OverFeat, a variant of convolutional neural networks (CNN), to do the defect classification. They have also shown that the fixed features from the pre-trained network perform well on some defects and perform poorly on texture kind of defect. These authors have also proposed a structural visual inspection method based on faster region-based CNN (faster R-CNN) to ensure quasi-real-time simultaneous detection of multiple types of defects \cite{cha2018autonomous}. \cite{song2019weak} proposed to detect weak scratches using deep CNN and skeleton extraction.  They have shown that the their method is robust to background noise. \cite{li2018real} proposed a variant of  you only look once (YOLO) network to detect surface defects of flat steel in real-time. The CNN is used extensively for a different kind of defect classification on different data sets \cite{zhou2017classification,yi2017end}. \cite{natarajan2017convolutional} use a transfer learning approach for defect classification. They have shown that transfer learning can help achieve a good classification accuracy with fewer data samples. \cite{ren2017generic} use a patch-wise classification to do both defect classification and segmentation. 

There are various methods in the literature on defect localization. \cite{soukup2014convolutional} uses a classical CNN to perform steel defect detection. Authors have explored the effect of regularization and unsupervised pre-training. \cite{he2019end} uses a pre-trained ResNet to extract the multi-scale features, and the features from different scales are fused using a multilevel feature fusion network (MFN). The fused features and region proposal network are used to classify the defect type and predict the bounding box. The main drawback of the method is that localization is very coarse. \cite{amin2020deep} use an U-Net and residual U-Net architecture for the fine segmentation of the steel defect. The method's main drawback is that the networks are trained with random initialization and need a large amount of pixel-level annotation of the defects. The pixel-level annotation process can be very time consuming and expensive. \cite{tabernik2020segmentation} uses a SegNet based semantic segmentation for the steel defect detection. There are various unsupervised and reinforcement learning based methods for steel defect detection. The summary of various methods for defect classification, and segmentation can be found in  \cite{luo2020automated}. The authors discuss the taxonomy of defect detection including the  statistical, spectral, model based and machine learning methods.

In this work, we systematically study the transfer learning effectiveness for steel defect classification and localization (SDCL). The transfer learning or domain adaption aims to reuse the feature learned in one domain to improve the learning in the other domain. This is a popular approach in cases where the annotated data is limited. The transfer learning has shown really good application in various tasks such a object detection \cite{talukdar2018transfer}, semantic segmentation \cite{belagali2020two,sun2019not} etc. It is already shown that the transfer-learning from an arbitrary domain to another domain may not be useful. Transfer learning is most effective when two domains are similar \cite{sun2019not,david2010impossibility}. Hence, it is important to study the effectiveness of transfer learning in the case of SDCL. We consider a baseline architecture of U-Net for steel defect segmentation. U-Net has demonstrated state of the art performance in various image segmentation tasks \cite{ronneberger2015u}. It uses an encoder-decoder architecture with skip connections. The encoder learns the images' features at different scales, and the decoder uses these features to predict the segmentation masks. In this work, we explore two kinds of pre-trained encoder networks-- ResNet and DenseNet networks. Both of these networks have been shown to perform well on various computer vision tasks. The networks are pre-trained on the ImageNet data set \cite{deng2009imagenet}. We use a linear classifier using the bottleneck representation of U-Net to classify the defect. We fine-tune both the encoder and decoder of the network using the Severstal dataset \cite{sevdataset}. The experiments on Severstal data shows that the performance of both segmentation and classification is better in case of the pre-trained network compared to the random initialization. It is found that  performance gain by using the pre-trained networks is even higher if 50\% of data is used for training. We also show that the convergence of transfer learning is faster compared to random initialization.

\section{Proposed Transfer Learning based U-Net}
The proposed architecture for joint steel defect segmentation and classification is shown in Fig. \ref{fig:pa_arch}. The architecture takes an input image of dimension $H\times W$ and classifies each pixel to be one or more type of the defect. It involves mainly four parts (1) The U-Net architecture, (2) the type of initialization (3) classification and (4) objective function 
\subsection{U-net architecture} 
The U-Net is an encoder-decoder architecture with a skip connection. The encoder encodes the image using an encoder block and reduces the resolution using pooling. This helps in extracting a multi-scale feature of images. The decoder up-samples the representation in every step. The skip connection can enable the decoder to select the feature at a different scale to make a more accurate prediction of the object boundaries. The output of the U-Net is $256\times 1600\times N$ with sigmoid activation, where N is the types of steel defects. 

\subsection{Transfer learning}
We explore two kinds of encoder blocks for transfer learning. Both of these nets are trained using ImageNet data set \cite{deng2009imagenet}. We briefly review the features of the two networks in the following subsections.

\begin{figure}
    \centering
    \includegraphics[scale=0.65,clip=true]{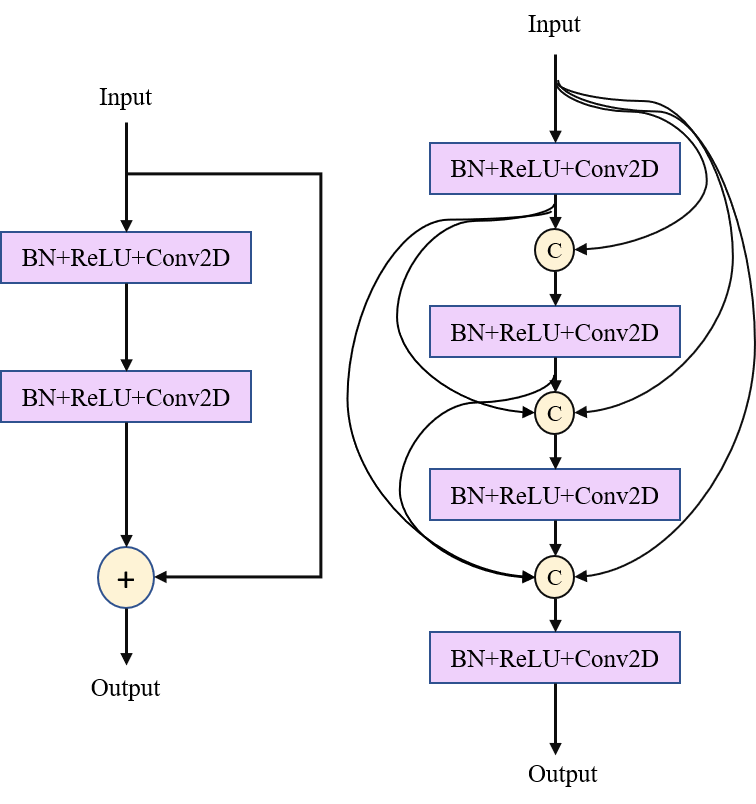}
    \caption{The structure of encoder layer Resnet (left) and the Densenet (right). The concatenation of inputs is indicated by (c) and + indicate the add operation. BN+ReLU+Conv2D indicate the batch normalization, Relu activation and convolution of kernel size 3x3 }
    \label{fig:resnet_densenet}
\end{figure}

\subsubsection{ResNet}
The residual networks (ResNet) are very deep convolutions neural networks with skip connection \cite{he2016deep}. The vanishing gradient problem is addressed by having a skip connection after each block. Each block contains a two 3x3 convolution with batch normalization and ReLU activation. Fig \ref{fig:resnet_densenet} (left) shows one layer of ResNet. The total parameters of the encoder are 11 million. 

\subsubsection{Densenet-121}
Densely connected convolutions neural nets (DenseNets) are stacked convolution networks where the feature map of the $L$-th layer is concatenated with the feature maps of the previous layer \cite{huang2017densely}. This has been shown to alleviate the vanishing gradient problem. The network's representation power is also increased because the deep layer has access to the previous layer feature maps. Fig \ref{fig:resnet_densenet} (right) shows one layer of DenseNet. The total parameters of this encoder are 6 million.

\subsection{Classification}
The encoder output encodes the rich abstract representation of the input image. Hence we propose to spatial average pooling the encoder output to extract the image representation. The image representation is passed through a linear classifier with sigmoid activation to enable the multi-label classification. 

\subsection{Objective function}
The joint segmentation and classification problem is formulated as a weighted combination of the two losses, as shown below.
\begin{equation}
\resizebox{.9\hsize}{!}{$\mathcal{J}= \sum_{k=1}^L\sum_{m=1}^N \left[ BCE(\hat{y}_{km},{y}_{km})+\sum_{j=1}^{256} \sum_{i=1}^{1600}  BCE(z_{kijm},\hat{z}_{kijm})\right]$}
\label{obj_fun}
\end{equation}
where $BCE$ indicate the binary cross-entropy loss, $k$ indicate the data point index, $m$ is the defect class index, $i,j$ are the spatial index, $\hat{y}$ indicate the predicted probability and $y$ is the ground truth defect labels. The predicted and ground truth pixel labels are indicated by $z$ and $\hat{z}$. During the test stage, the labels from the probability are obtained using the threshold of 0.5.

\section{Experiments and results}
\subsection{Data-set and Pre-processing}

The Kaggle Competition - "Severstal: Steel Defect Detection`` data is used for all the experiments. In each experiment, the input image could contain one or more kinds of defects. The training set includes 12568 images, and 6666 of them include at least one defective region. Ground truth classification was performed by an expert to provide the defect type classification and the annotation of the defective region by visual inspection. The resolution of images is 256x1600 px. We normalize the image using the global mean and standard deviation. We apply a random vertical/horizontal flip as data augmentation. The same augmentation applies to both the original image and the corresponding ground truth masks to pair with the augmented images.

\begin{figure*}
    \centering
    \includegraphics[width=0.9\textwidth,trim={4cm 0cm 0 0},clip=true]{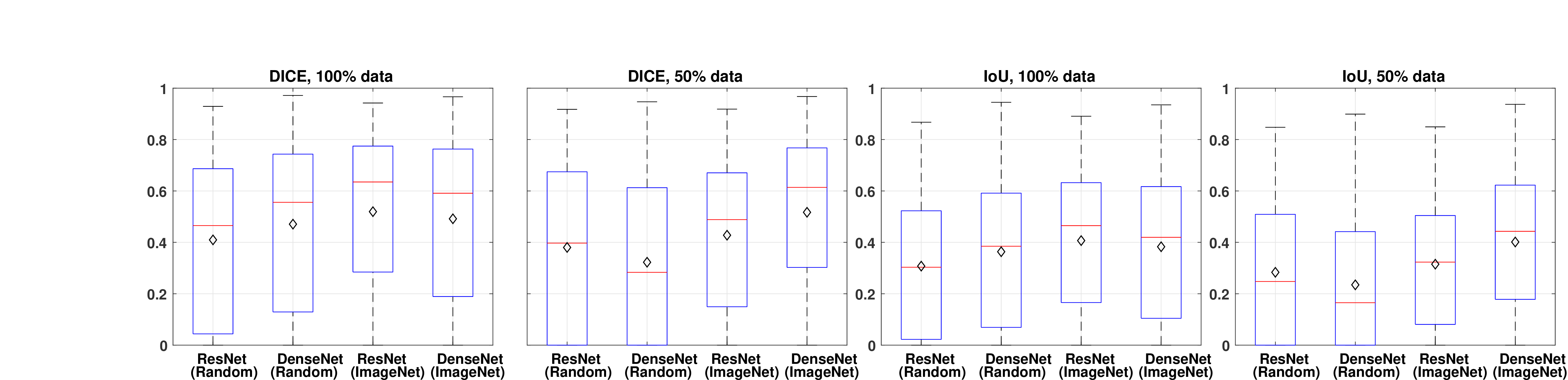}
    \caption{Boxplot comparison of DICE and IoU for different networks and the initialization. The red line indicate the median, the black dot indicate the mean, the blue box indicate the 75\% confidence interval and the black whiskers indicate the 95\% confidence intervals.}
    \label{fig:seg_res}
\end{figure*}

\subsection{Experimental setup}
We use a total of 75\% data for the training, 12.5 \% data for validation, and 12.5\% data for the testing. The U-Net of five encoders and decoder is used for all the experiments. The network is trained using the objective function in eq. \ref{obj_fun} with the batch size 16 and Adam optimizer with learning rate of $5\times 10^{-4}$, $\beta_1$=0.99 and $\beta_2$=0.99 \cite{kingma2014adam}. We train the network for 10 epochs with early stopping. We use the U-Net with random initialization as the baseline.  We have implemented the network in PyTorch \cite{paszke2017automatic} with the PyTorch segmentation library \cite{Yakubovskiy:2019}. The ResNet/DenseNet with random initialization is indicated by ResNet(Random)/DenseNet(Random). The Imagenet pre-trained counterparts are indicated by ResNet(Imagenet)/DenseNet(Imagenet) respectively. To understand the model's sample complexity, we also train these networks using 50\% of the training data. We refer to the pre-trained initialization as TLU-Net and the random initialization as just U-net.

\subsection{Evaluation metrics}
We evaluate the steel defect classification performance using multi-label classification accuracy (MLA) and the average area under receiver operating curve (AUC) across 4 classes. The MLA is defined as the proportion of the predicted correct labels to a total number of labels for instance. We treat the multi-label classification as 4 separate binary classifiers and compute the average AUC  across four classifiers. 

We have used DICE and the intersection of unions (IoU) to evaluate the performance of steel defect segmentation. The DICE metric for each class is defined as follows:
\begin{equation}
DICE=\frac{2|X\cap Y|}{|X|+|Y|}    
\end{equation}
The IoU is defined as follows:

\begin{equation}
IoU=\frac{|X\cap Y|}{|X \cup Y|}
\end{equation}

where $X$ and $Y$ are the ground-truth and the predicted segmentation masks, $\cup$ indicates the union operation, $\cap$ indicated the intersection, and $|.|$ indicates the cardinally. 

\begin{figure}
    \centering
    \includegraphics[width=\columnwidth,trim={0cm 0cm 0 0},clip=true]{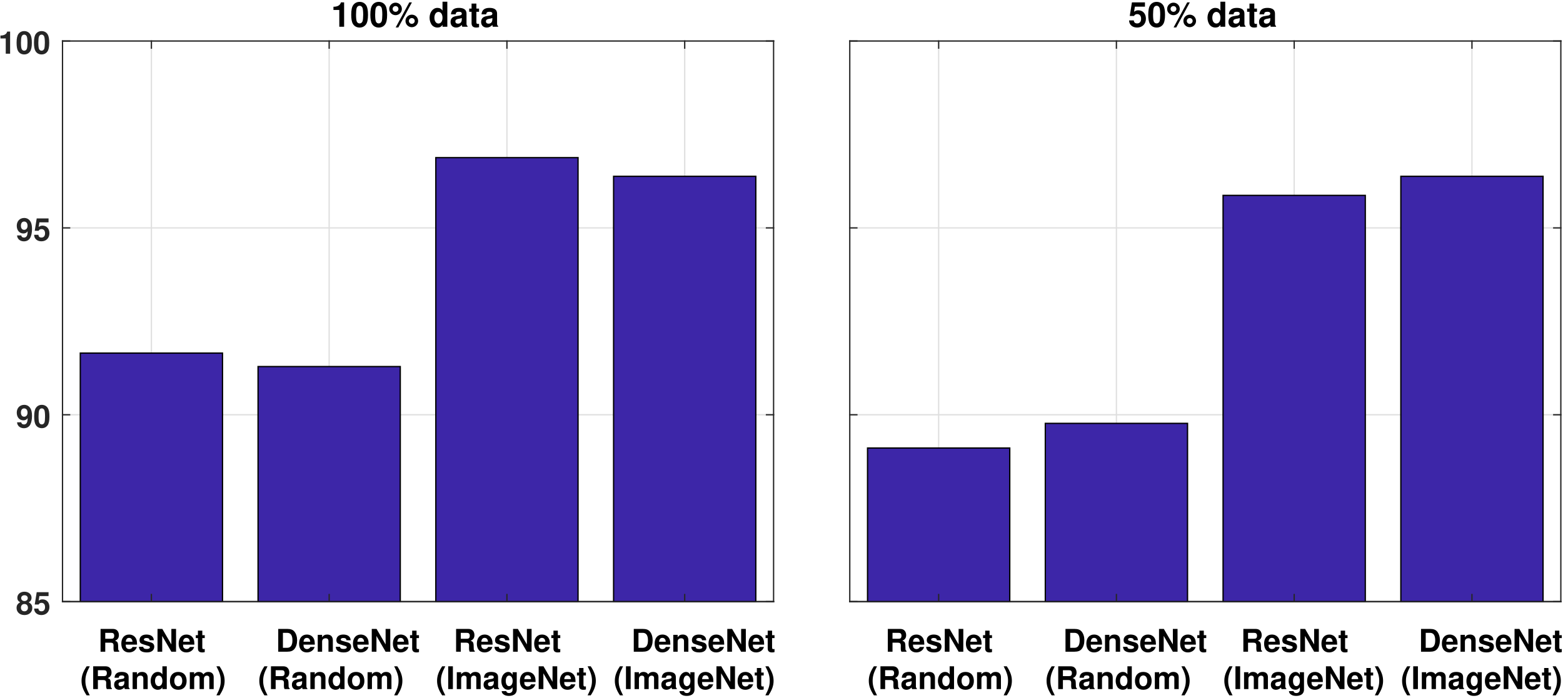}
    \caption{Comparison of average MLA for different networks and initialization using different amount of training data.}
    \label{fig:acc}
\end{figure}

\subsection{Results and discussion}
Fig. \ref{fig:acc} shows the MLA(\%) comparison for different networks and the initialization. It is clear from the figure that the TLU-Net achieves $~5\%$ (absolute) improvement in the  MLA compared to the random initialization with 100\% training data. This indicates that the features learned using ImageNet can help for steel defect classification as well.   The MLA gap increases to $~8\%$ (absolute) as the training data is reduced to $50\%$. The performance of TLU-Net does not drop significantly as the training data is reduced.  This indicates that the TLU-Net is helpful when there is a limited number of annotated data points. The ResNet(ImageNet) and DenseNet(ImageNet) performs best in case of 100\%  and 50\% of training data respectively. This could be because the DenseNet has fewer parameters than the ResNet and hence needs less data but has less representational power.
\begin{figure}
    \centering
    \includegraphics[width=\columnwidth,trim={0cm 0cm 0 0},clip=true]{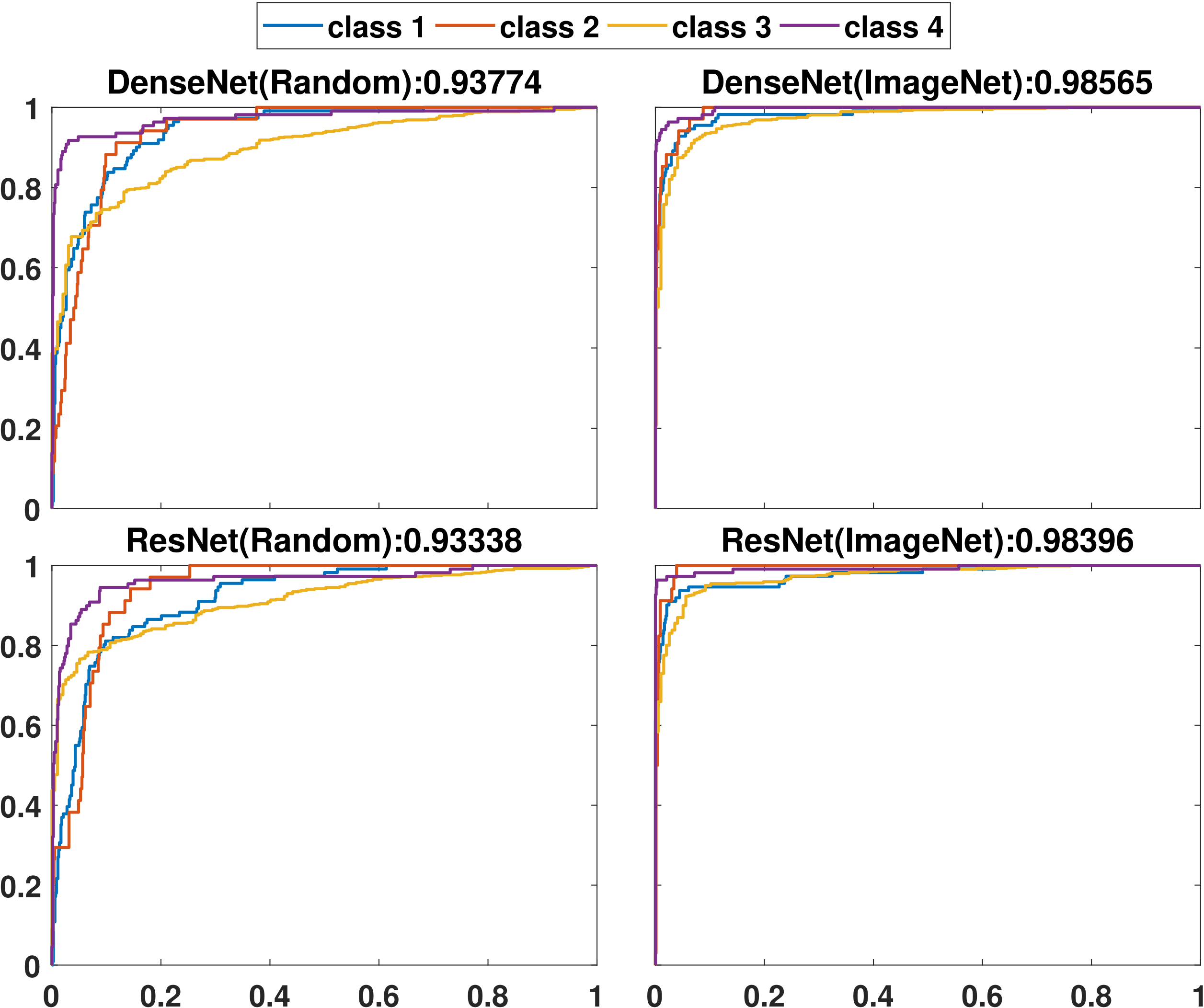}
    \caption{comparison of AUC plot of four classes for different networks and initialization. The title shows the average AUC across four classes. }
    \label{fig:auc}
\end{figure}

Fig. \ref{fig:auc} shows the AUC for different networks and initialization. It is clear from the figure that the best AUC archived for class 1 in all cases. Class 3 and class 4 defects demonstrate the poorest performace. This is mainly because of the number of samples of the training data for each class. The AUC of all classes improved by using the TLU-Net compared to U-Net. The DeseNet(ImageNet) has the highest AUC. 

\begin{figure*}
    \centering
    \includegraphics[width=0.95\textwidth,trim={4cm 5cm 4cm 0},clip=true]{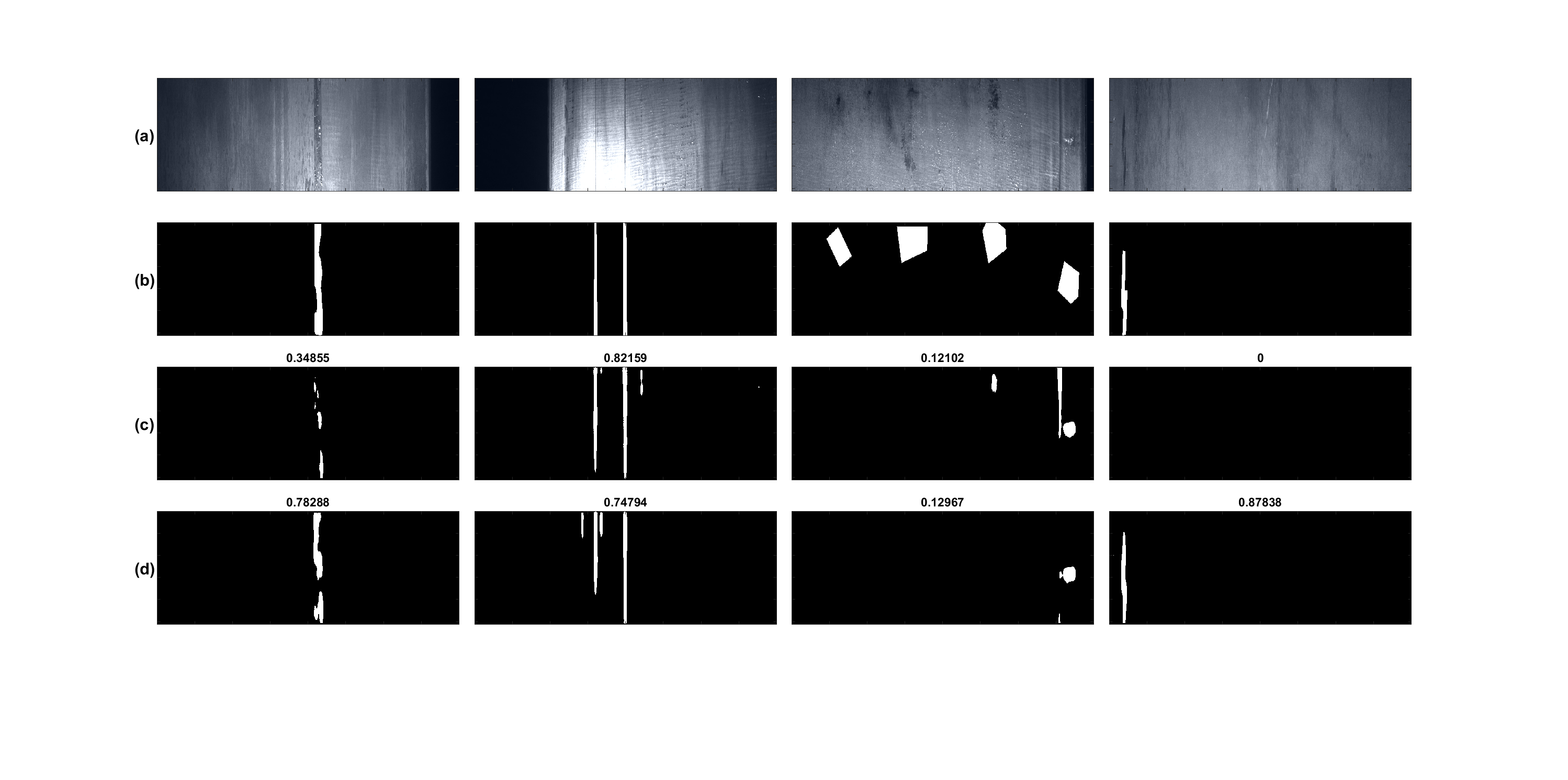}
    \caption{Illustration of segmentation mask prediction. (row a) The input images (row b) The ground truth masks (row c) The mask predicted by ResNet(Random) (row d) the mask predicted by ResNet(ImageNet). The corresponding DICE for the prediction is shown in the title of the image.}
    \label{fig:ex_fog}
\end{figure*}

Fig. \ref{fig:seg_res} shows the box plot of the DICE and IoU for different network with the 100\% and 50\% of the training data. In all cases, the median and mean performance of the TLU-Net perform better than that of the U-Net. In case of 100\% training data, the ResNet(ImageNet) DICE/IoU is better than all the other models and it has the small 75\% confidence interval.  The TLU-net with ResNet shows an improvement of $\sim$26\% (relative) compared to U-Net with ResNet. The TLU-net with DenseNet shows an improvement of $\sim$5\% (relative) compared to U-Net with ResNet.
But in the case of 50\% training data, the DenseNet with transfer learning shows an improvement of 60\%(relative), and the ResNet shows an improvement of 12\%(relative). This clearly indicate the gain of using transfer learning is higher as the number of annotated samples are lower. 

\begin{figure}
    \centering
    \includegraphics[width=\columnwidth,trim={0cm 0cm 0 0},clip=true]{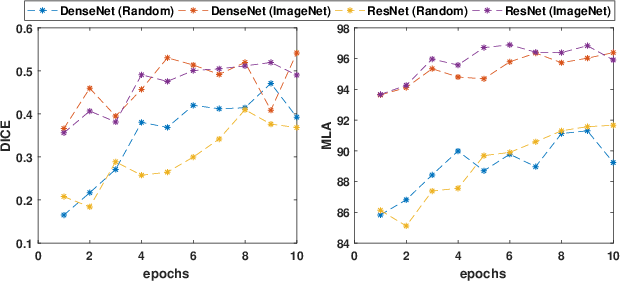}
    \caption{Comparison of validation loss/MLA evolution over epoch for different networks and initialization.}
    \label{fig:convergence}
\end{figure}

Fig. \ref{fig:convergence} shows the DICE/MLA metric using validation data during the course of training. This helps us understand the convergence rate of different networks. It is clear from the figure that the TLU-Net has higher DICE at the beginning of the epochs than the U-Net. The converged DICE value for the TLU-Net is higher compared to the U-Net. This clearly indicates that transfer learning is helping in faster convergence of the model. Similar observations are applicable for the MLA plot also. It is interesting to note that the pre-trained model's starting accuracy is significantly higher than the random initialization. This is mainly because the classifier directly uses the encoder output for classification, and the encoder is initialized with the pre-trained weights. It implies that the pre-trained features are useful in discriminating against the different steel defects.


\begin{figure}
    \centering
    \includegraphics[width=\columnwidth,trim={0cm 0cm 0 0},clip=true]{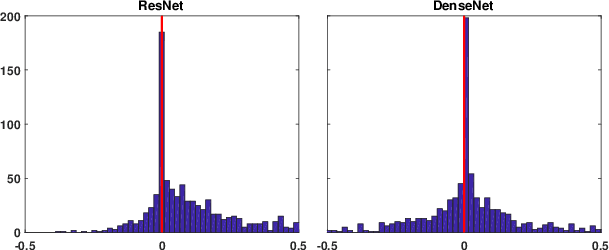}
    \caption{Histogram of difference between DICE obtained using transfer learning and the random initialization. The red line indicate the zero line. }
    \label{fig:hist_err}
\end{figure}

Fig. \ref{fig:ex_fog} shows the four illustrative examples of mask predictions for TLU-Net using ResNet and the U-Net using ResNet. In the first column, the TLU-Net can detect the defect more accurately, and the U-Net fails to detect some parts of the defects. In the second column, the TLU-net is showing some false positive detection because of some illumination differences. In the third column, both networks  failed to detect some of the defects. This mainly because the training data has few defects belonging to this class. We hypothesize that the TLU-net also fails because the defect share is more complex, and pre-learned features may not help this scenario. In the fourth column, the TLU-Net can detect the defect in the form of a fine line, and the U-Net is failing to detect these lines. 

Fig. \ref{fig:hist_err} shows the histogram of the DICE difference between the TLU-Net and the U-Net using both ResNet/DenseNet. It is clear from the figure that the histogram is skewed toward positive values. We observe improvement in the case of 81\% images in case ResNet and 63\% of images in DenseNet.

\section{Conclusion}

In this work, we propose to use the transfer learning framework for steel defect classification and segmentation. We use a U-Net architecture as a base architecture and explore two kinds of encoders: ResNet and Dense Net. We compare these nets' performance using random initialization and the pre-trained networks trained using ImageNet data set. We found that the performance of the transfer learning is superior both in terms of defect segmentation and classification. We also found the performance gap increases as the training data decreases. We also found that the convergence rate with transfer learning is better than that of the random initialization. We have found that transfer learning performance is poor in rare defect types and complex shape defects. As a part of future work, we would like to work on transfer learning to handle more complex shapes using synthetic data and the rare defect type generalization using generative models. We want to explore the semi/weakly supervised learning approaches to reduce the annotated training data requirement. 
\printbibliography
\end{document}